\documentclass[11pt]{article}

\usepackage[preprint]{acl}

\usepackage{times}
\usepackage{url}
\usepackage{latexsym}
\usepackage{amsmath} 
\usepackage{booktabs}
\usepackage{graphics}
\usepackage{graphicx}
\usepackage{multirow}
\usepackage[utf8]{inputenc}
\usepackage{CJKutf8} 
\usepackage{amssymb}
\usepackage{footmisc}
\usepackage{algorithm}
\usepackage{algorithmic}
\usepackage{xspace,mfirstuc,tabulary}
\usepackage{tabularx}
\usepackage{longtable}
\usepackage{arydshln} 
\usepackage{listings}

\newcommand{\zhname}[1]{\begin{CJK}{UTF8}{gbsn}#1\end{CJK}}
\newcommand\blfootnote[1]{%
  \begingroup
  \renewcommand\thefootnote{}\footnote{#1}%
  \addtocounter{footnote}{-1}%
  \endgroup
}
\usepackage[T1]{fontenc}

\usepackage[utf8]{inputenc}

\usepackage{microtype}

\usepackage{inconsolata}

\usepackage{graphicx}
\usepackage{hyperref}
\title{Is Human Annotation Necessary? Iterative MBR Distillation for\\Error Span Detection in Machine Translation}

\author{Boxuan Lyu \zhname{吕博轩}$^{\text{1}}$, Haiyue Song \zhname{宋海越}$^{\text{2}}$ and Zhi Qu \zhname{屈直}$^{\text{2}}$\\$^{\text{1}}$Institute of Science Tokyo, \\$^{\text{2}}$National Institute of Information and Communications Technology, Japan, \\\texttt{\url{lyu@lr.first.iir.isct.ac.jp}}\\
\texttt{\{\href{mailto:haiyue.song@nict.go.jp}{haiyue.song}, \href{mailto:qu.zhi@nict.go.jp}{qu.zhi}\}@nict.go.jp}}

\begin{document}
\maketitle
\begin{abstract}
Error Span Detection (ESD) is a crucial subtask in Machine Translation (MT) evaluation, aiming to identify the location and severity of translation errors. 
While fine-tuning models on human-annotated data improves ESD performance, acquiring such data is expensive and prone to inconsistencies among annotators. 
To address this, we propose a novel self-evolution framework based on Minimum Bayes Risk (MBR) decoding, named \textbf{Iterative MBR Distillation for ESD}, which eliminates the reliance on human annotations by leveraging an off-the-shelf LLM to generate pseudo-labels.
Extensive experiments on the WMT Metrics Shared Task datasets demonstrate that models trained solely on these self-generated pseudo-labels outperform both unadapted base model and supervised baselines trained on human annotations at the system and span levels, while maintaining competitive sentence-level performance.\blfootnote{This  work was done during the first author's internship at National Institute of Information and Communications Technology, Kyoto, Japan.}\footnote{Our models and code are available at: \url{https://github.com/vlaks425/MBR-ESD}}
\end{abstract}

\section{Introduction}
In the landscape of Machine Translation (MT) evaluation, Error Span Detection (ESD) plays a pivotal role. By identifying not only the presence of errors but also their precise locations and severity, ESD provides the granular feedback essential for MT model correction \cite{wmtqe23,wmtm24,wmtm25}.

However, despite its critical importance, the advancement of ESD models is severely constrained by major bottlenecks in supervision.
Unlike sentence-level quality estimation, ESD demands fine-grained span-level annotation, which is a costly process that requires bilingual expertise \cite{mqm,esa}.
Furthermore, the gold-standard status of human labels is challenged by inherent subjectivity. 
As noted by Proietti et al.~\shortcite{proietti}, under existing annotation protocols, the agreement among human annotators is merely comparable to the agreement between automatic annotators and humans.
Consequently, public ESD datasets remain limited in scale and consistency compared to the massive corpora available for general MT tasks.
This data scarcity and inherent subjectivity limit the potential of supervised ESD models \cite{metricx25} and prompt a fundamental question:
\textit{Is human annotation strictly necessary to train ESD models?}

\begin{table}[t]
\centering
\begin{tabular}{lccc}
\hline
\textbf{Method} & SPA & ${\text{Acc}}_{\text{eq}}^{*}$ & \textsc{SoftF1} \\
\hline
Base Model & .851& .518 & .874 \\
Gold-SFT & .826 & .571 & .915 \\
\textbf{MBR Distill} & \textbf{.864} & \textbf{.573} & \textbf{.933} \\
\hline
\end{tabular}
\caption{Our self-trained MBR Distillation framework surpasses both base model and human-annotated data trained baseline (Gold-SFT) at the system (SPA) and span (\textsc{SoftF1}) levels.}
\label{tab:highlight}
\end{table}
In this work, we challenge the reliance on human annotation data by leveraging purely synthetic training data generated through \textbf{Iterative Minimum Bayes Risk (MBR) Distillation for ESD}, a novel framework for ESD model training.
Inspired by the recent success of Large Language Models (LLMs) in self-evolution \cite{surveyselfevolution}, we hypothesize that LLMs, possessing latent evaluation capabilities, can synthesize their own training signals.
To construct pseudo-labels, we employ MBR decoding \cite{mbr_speech1,mbr_speech2}.
By selecting candidates that minimize the expected risk across a diverse sample set, this approach effectively leverages the model's internal ``consensus'' to filter out noise \cite{esd-mbr}.

Our approach operates as an iterative cycle.
Starting with an off-the-shelf LLM, we generate diverse candidate error spans and select pseudo-labels via MBR decoding with the \textsc{SoftF1} utility function \cite{esd-mbr}.
The model is then fine-tuned on these self-generated labels using diverse training objectives, including Supervised Fine-Tuning (SFT), Direct Preference Optimization (DPO) \cite{dpo}, and Kahneman-Tversky Optimization (KTO) \cite{kto}.
Through iterations of this cycle, the model drives its own continuous improvement.

Our experiments yield a compelling and counter-intuitive finding: models trained exclusively on MBR-generated pseudo-labels outperform those fine-tuned on human annotations at both the system and span levels, while remaining competitive at the sentence level (as highlighted in Table~\ref{tab:highlight}).

The main contributions of this paper are summarized as follows:
\begin{itemize}
    \item We propose Iterative MBR Distillation for ESD, a novel self-evolution framework that completely bypasses the need for human annotations.
    \item We provide a comprehensive evaluation across various training objectives (SFT, DPO, KTO). Empirical results demonstrate that our framework not only surpasses unadapted base model but also outperforms models trained on human annotations at the system and span levels, signaling a paradigm shift in ESD model training.
\end{itemize}

\section{Related Work}
\label{sec:rw}

\subsection{MT Automatic Metrics}
\label{sec:rw:metric}
Early MT metrics such as BLEU \cite{bleu} and chrF \cite{chrf} rely on surface-level matching and output a single scalar score per sentence. 
Neural metrics, including COMET \cite{comet} and MetricX \cite{metricx25}, show higher correlations with human judgments but typically still produce only sentence-level scores. 
As human evaluation has shifted toward multidimensional annotations, such as Multidimensional Quality Metrics (MQM) \cite{mqm}, automatic metrics have targeted finer-grained evaluation. 
Generative approaches, such as GEMBA-MQM \cite{gemba_mqm}, prompt LLMs to output MQM-style error spans.

\subsection{Self-Evolution of LLMs}
\label{sec:rw:self-evo}
Self-evolution empowers LLMs to autonomously acquire and refine knowledge by synthesizing training data and learning from their own generated feedback \cite{surveyselfevolution}. 
In the context of machine translation, recent studies have explored self-evolutionary frameworks to continuously enhance translation capabilities. 
For example, methods like self-evolution knowledge distillation allow models to dynamically assess token-level learning difficulties and adjust prior knowledge during the training process \cite{mt_se}. 
Similarly, iterative self-correction frameworks leverage the LLM's own estimations to refine initial translations, demonstrating that models can bootstrap their performance without relying heavily on external human supervision \cite{mt_sr}.

While self-evolution has shown significant promise in text generation and translation, most existing works do not address the self-evolution of ESD models. 
The only closely related study is by Lyu et al.~\shortcite{esd-mbr} who introduced MBR decoding for generative ESD models. 
To alleviate computational bottlenecks, they investigated MBR distillation to transfer the superior performance of MBR decoding into the model weights. 
They employed DPO to train the model to approximate MBR outputs, confirming that the distilled model using greedy search can match the performance of full MBR decoding. 
However, their exploration has two key limitations: 
(1) it did not include a performance comparison against models trained on human annotations, and (2) it did not propose an iterative training mechanism. 
Our work bridges these gaps by introducing Iterative MBR Distillation for ESD, demonstrating that a self-evolving iterative framework can effectively surpass the performance of models fine-tuned on human-annotated data.

\begin{algorithm}[p]
\caption{Iterative MBR Distillation for ESD}
\label{alg:iter_mbr}
\begin{algorithmic}[1]
\REQUIRE Unlabeled data $\mathcal{D}_{u}=\{x_i\}_{i=1}^{N}$; base model $M_{\theta^{(0)}}$; iterations $T$; candidate count $C$; utility $u=\textsc{SoftF1}$; loss function $\mathcal{L}$
\FOR{$t=1$ to $T$}
  \STATE $\mathcal{D}^{(t)} \leftarrow \emptyset$
  \FOR{each $x_i \in \mathcal{D}_{u}$}
    \STATE Sample $\mathcal{C}_{i}^{(t)}$ ($|\mathcal{C}_{i}^{(t)}|=C$) from $M_{\theta^{(t-1)}}(\cdot \mid x_i)$
    \STATE Set $\mathcal{S}_{i}^{(t)} \leftarrow \mathcal{C}_{i}^{(t)}$
    \STATE Calculate $score_{E}^{MBR}$ for all $E^c \in \mathcal{C}_{i}^{(t)}$ using Eq.~\ref{def:appox_mbr}
    \STATE $E_i^{+} \leftarrow \operatorname*{argmax}_{E \in \mathcal{C}_{i}^{(t)}} score_{E}^{MBR}$
    \STATE $E_i^{-} \leftarrow \operatorname*{argmin}_{E \in \mathcal{C}_{i}^{(t)}} score_{E}^{MBR}$
    
    \IF{$\mathcal{L}$ is $\mathcal{L}_{\text{SFT}}$}
      \STATE Add $(x_i, E_i^{+})$ to $\mathcal{D}^{(t)}$
    \ELSIF{$\mathcal{L}$ is $\mathcal{L}_{\text{DPO}}$ \AND $E_i^{+} \neq E_i^{-}$}
      \STATE Add $(x_i, E_i^{+}, E_i^{-})$ to $\mathcal{D}^{(t)}$
    \ELSIF{$\mathcal{L}$ is $\mathcal{L}_{\text{KTO}}$}
      \STATE Add $(x_i, E_i^{+})$ as $E^+$ to $\mathcal{D}^{(t)}$
      \IF{$E_i^{+} \neq E_i^{-}$}
        \STATE Add $(x_i, E_i^{-})$ as $E^-$ to $\mathcal{D}^{(t)}$
      \ENDIF
    \ENDIF
  \ENDFOR
  \STATE Update $\theta^{(t)}$ by minimizing $\mathcal{L}(\theta^{(t)})$ on $\mathcal{D}^{(t)}$
\ENDFOR
\RETURN $M_{\theta^{(T)}}$
\end{algorithmic}
\end{algorithm}

\begin{figure*}
    \centering
    \includegraphics[width=0.9\linewidth]{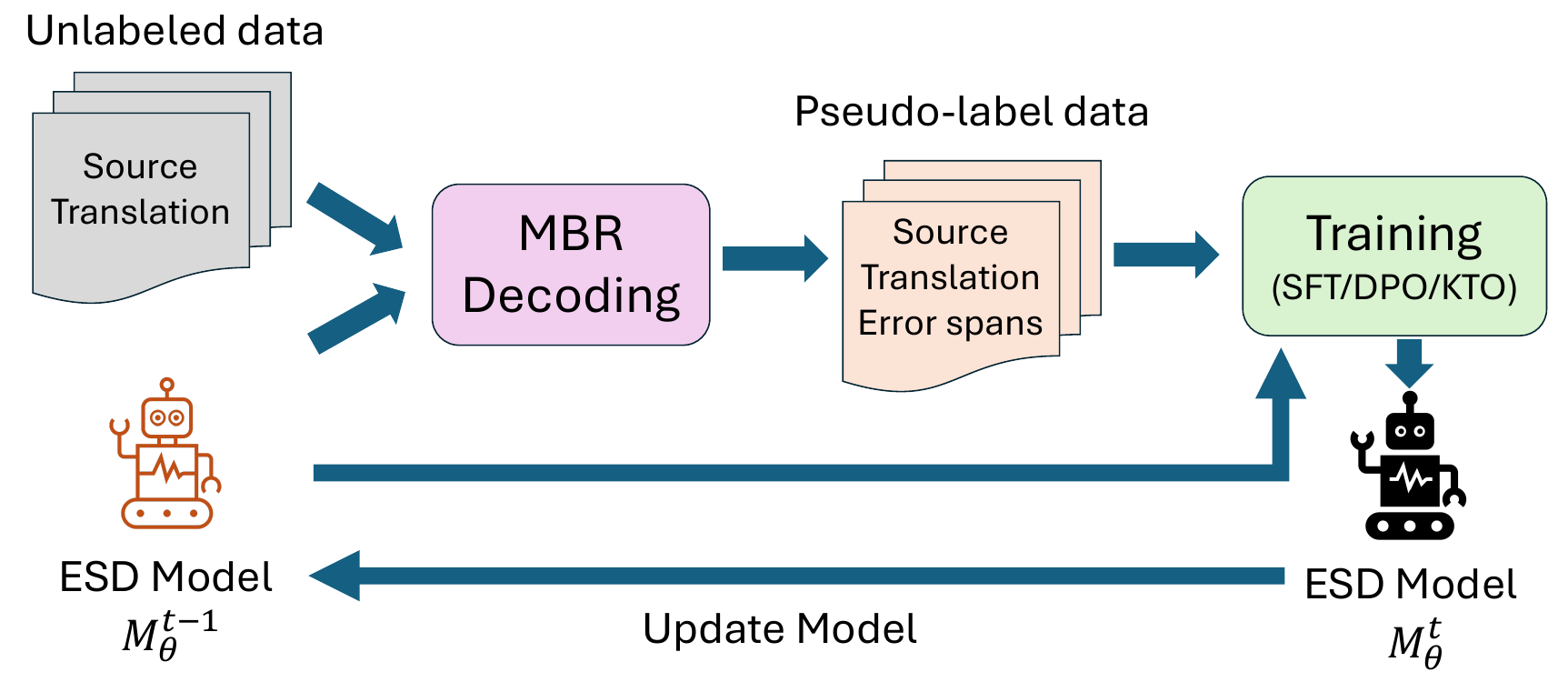}
    \caption{Overview of the Iterative MBR Distillation framework for ESD. Starting with unlabeled source-translation pairs, the model generates diverse candidate error spans. MBR decoding then evaluates these candidates to assign utility scores, identifying high-quality pseudo-labels (e.g., the best and worst hypotheses). Finally, the model is fine-tuned on these self-generated labels using SFT, DPO, or KTO. This cycle repeats iteratively, enabling the model to self-evolve and refine its ESD capabilities without relying on human annotations.}
    \label{fig:placeholder}
\end{figure*}
\section{Preliminaries}

\subsection{Error Span Detection Model}
Formally, let $x = (s, t)$ denote an input pair, where $s$ is a source sentence and $t$ is its corresponding translation. 
The ESD model, denoted as $M_{\theta}$ with parameters $\theta$, generates an output $E$. 
Here, $E$ represents a set of error annotations, specifically the character positions and severity values for each error span detected in $t$.

\subsection{MBR Decoding for ESD}
MBR decoding has been extensively utilized in sequence generation tasks, such as automatic speech recognition \cite{mbr_speech1,mbr_speech2} and MT \cite{mbr_nmt1,mbr_nmt2} to address the shortcomings of Maximum a Posteriori (MAP) decoding.
While MAP relies on the assumption that the model's probability perfectly aligns with generation quality, MBR aims to minimize the expected risk, or equivalently, maximize the expected utility, defined by a task-specific evaluation metric, e.g., BLEU \cite{bleu} for MT.

Recently, Lyu et al.~\shortcite{esd-mbr} proposed applying MBR decoding to generative ESD models.
By selecting the hypothesis (error annotation) that maximizes utility against a set of samples drawn from the model, MBR allows ESD models to effectively aggregate consensus and mitigate prediction errors inherent in standard MAP decoding.

Formally, let $\mathcal{C} = \{E_1^{c}, \dots, E_N^{c}\}$ be the set of candidate hypotheses.
MBR decoding first assigns a $score_{E}^{MBR}$ to each candidate $E^c$, and then selects the hypothesis $E^{\mathit{MBR}}$ with the highest score:
\begin{align}
E^{\mathit{MBR}} & = \mathop{\mathrm{argmax}}score_{E}^{MBR}\nonumber,
\end{align}
where $score_{E}^{MBR}$ is defined as the expected utility with respect to the human annotation distribution $\mathcal{Y}$:
\begin{align}
score_{E}^{MBR}
& =
{\mathbb{E}_{y \in \mathcal{Y}}\left[u(E^c, y) \mid x\right]}\nonumber.
\end{align}
Since the human annotation is unavailable during inference, this expectation is approximated using a finite set of support hypotheses $\mathcal{S}$ sampled from the model.
Formally, for a given input, we generate a support set $\mathcal{S}$ (where typically $\mathcal{S} = \mathcal{C}$) from the model $M_{\theta}(\cdot|x)$.

If we assume that the probability of each support hypothesis is equal, and do not use the model-estimated probabilities (which is consistent with the fundamental assumption of MBR, that the model-estimated probabilities are not reliable), then the MBR score of the candidate $E^c$ can be approximated by its average utility to the support hypotheses:
\begin{align}
    score_{E}^{MBR} \approx \frac{1}{|\mathcal{S}|} \sum_{E^s \in \mathcal{S}} u(E^c, E^s),
    \label{def:appox_mbr}
\end{align}
where $u(\cdot, \cdot)$ is the utility function.

\subsection{ESD Model Training}
\label{sec:esd_train}

While LLM-based ESD models can perform inference without ESD-specific training \cite{gemba_mqm}, fine-tuning on ESD datasets has proven to improve model performance \cite{metricx25}.
The goal of training an ESD model is to tune its parameters to make its predictions more closely resemble the pre-defined target annotations (e.g., human annotations).

Formally, we assume access to a labeled dataset $\mathcal{D} = \{(x_i, E^t_i)\}_{i=1}^N$, where $E^t_i$ denotes the target annotation.
The general training objective is to find the optimal parameters $\theta^*$ that minimize a specific loss function $\mathcal{L}$ over the data distribution:
\begin{equation}
    \theta^* = \operatorname*{arg\,min}_\theta \mathbb{E}_{(x, E^t) \sim \mathcal{D}} [\mathcal{L}(\theta; x, E^t)].\nonumber
\end{equation}
This general objective can be optimized through different paradigms. 
For instance, we can employ SFT to perform maximum likelihood estimation directly on the target annotation, or utilize preference learning, such as DPO and KTO.

\paragraph{Supervised Fine-Tuning (SFT)}
SFT is the standard approach for alignment which maximizes the conditional log-likelihood of $E^t_{i}$:
\begin{equation}
  \mathcal{L}_{\text{SFT}}(\theta)
  = - \sum_{i} \log M_{\theta}(E^t_i \mid x_i).\nonumber
\end{equation}

\paragraph{Direct Preference Optimization (DPO)}
DPO \cite{dpo} is a pairwise preference optimization method designed to align language models with human feedback. 
Its primary optimization objective is to increase the margin of the estimated probabilities between the preferred and dispreferred outputs.
Let $M_{\theta}$ denote the policy being optimized and $M_{\text{ref}}$ the frozen reference policy.
Given a preference pair $(E_i^{+}, E_i^{-})$ where $E_i^{+} \neq E_i^{-}$ and the hypothesis $E_i^{+}$ is preferred over $E_i^{-}$, DPO optimizes:
\begin{align}
    \mathcal{L}_{\text{DPO}}(\theta)
= &- \sum_{i} \log \sigma\!(
  \lambda [
  \log \frac{M_\theta(E_i^+ \mid x_i)}{M_\theta(E_i^- \mid x_i)} \nonumber
   \\ 
   &-\log \frac{M_{\text{ref}}(E_i^+ \mid x_i)}{M_{\text{ref}}(E_i^- \mid x_i)}\nonumber
  ]),
\end{align}
where $\lambda$ is the regularization parameter (referred to as $\beta$ in standard DPO paper \cite{dpo}, but distinguished here to avoid confusion with KTO hyperparameters).

\paragraph{Kahneman-Tversky Optimization (KTO)}
KTO is a training objective designed to handle binary feedback signals, eliminating the strict requirement for paired preference data \cite{kto}.
While DPO dictates that each input $x$ must be accompanied by a paired $(E^+, E^-)$ such that $E^+ \neq E^-$, KTO evaluates examples individually. It requires only that at least one of $E^+$ (desirable) or $E^-$ (undesirable) is present for a given $x$, although both can be utilized if available.
Following Ethayarajh et al.~\shortcite{kto}, the loss function is defined as:
\begin{equation}
    \mathcal{L}_{\text{KTO}}(\theta) = \mathbb{E}_{(x, E) \sim \mathcal{D}} [ w_E (1 - \sigma(v(x, E))) ],\nonumber
\end{equation}
where $E \in \{E^+, E^-\}$ and $w_E$ is the corresponding weight hyperparameter depending on whether $E$ is desirable ($E^+$) or undesirable ($E^-$).
The value function $v(x, E)$ is defined piecewise to reflect the asymmetry of the Kahneman-Tversky value model:
\begin{equation}
    v(x, E) = 
    \begin{cases} 
    \beta (r_{\theta}(x, E) - z_{\text{ref}}) & \text{if } E = E^+ \\
    \beta (z_{\text{ref}} - r_{\theta}(x, E)) & \text{if } E = E^-
    \end{cases}\nonumber
\end{equation}
Here, the implicit reward $r_{\theta}$ is the log-ratio between the policy and reference model:
\begin{equation}
    r_{\theta}(x, E) = \log \frac{M_\theta(E \mid x)}{M_{\text{ref}}(E \mid x)},\nonumber
\end{equation}
and $z_{\text{ref}}$ is a reference point (typically the Kullback--Leibler (KL) divergence) used to control the saturation of the loss.

\section{Proposed Method: Iterative MBR Distillation for ESD}
\label{sec:mbr_distill}
To combine the performance benefits of MBR decoding with the inference efficiency of direct generation, we propose an iterative self-training framework, as illustrated in Figure~\ref{fig:placeholder}. 
The core idea is to distill the knowledge from the MBR scoring process back into the model parameters.
As shown in Algorithm~\ref{alg:iter_mbr}, the method iterates through cycles of data generation and model updates. 
In each iteration, the model generates candidates for unlabeled data, assigns MBR scores to identify the best (and potentially worst) hypotheses, and constructs a synthetic training set. 
The model is then updated using one of the training objectives described in Section~\ref{sec:esd_train} (SFT, DPO, or KTO), effectively amortizing the cost of MBR decoding.

\section{Experiments}
\label{sec:exp}
\begin{table*}[t]
\centering
\begin{tabular}{lcccc}
\hline
\textbf{Method} & ~~~~SPA & ~~~~${\text{Acc}}_{\text{eq}}^{*}$ & \textsc{SoftF1} & ~\textsc{F1}\\
\hline
xCOMET-ESD &~~~.757&~~.553&.889&.302\\
xCOMET-Reg &~~~.844&~~\textbf{.581}&-&- \\
xCOMET-QE-ESD &~~~.688&~~.541&.879&.289\\
xCOMET-QE-Reg &~~~.825&~~.549&-&- \\
\hline
Base Model & ~~~.851 & ~~.518 & .874 & .396 \\
Gold-SFT & ~~~.826 & ~~.571 & .915 & .454 \\
Gold-DPO & ~~~.673 & ~~.518 & .909 & .472\\
Gold-KTO & ~~~.689 & ~~.567 & .910 & .469 \\
\hline
MBR Distill ($T=1$, SFT) & ~~~.851 & ~~.578 & .920 & .497 \\
MBR Distill ($T=2$, SFT) & ~~~.862 & ~~.564 & ~~\textbf{.939}\textsuperscript{†} & ~~.560\textsuperscript{†} \\
MBR Distill ($T=3$, SFT) & ~~~.857 & ~~.530 & ~~\textbf{.939}\textsuperscript{†} & ~~\textbf{.577}\textsuperscript{†} \\
\cdashline{1-5}
MBR Distill ($T=1$, DPO) & ~~~.843 & ~~.569 & .928 & ~~.528\textsuperscript{†}  \\
MBR Distill ($T=2$, DPO) & ~~~.860 & ~~.571 & ~~.932\textsuperscript{†} & ~~.539\textsuperscript{†} \\
MBR Distill ($T=3$, DPO) & ~~~.849 & ~~.572 & ~~.933\textsuperscript{†} & ~~.538\textsuperscript{†} \\
\cdashline{1-5}
MBR Distill ($T=1$, KTO) & ~~~.842 & ~~.573 & .925 & .529  \\
MBR Distill ($T=2$, KTO) & ~~~~~\textbf{.864}\textsuperscript{†} & ~~.573 & ~~.933\textsuperscript{†} & ~~.535\textsuperscript{†} \\
MBR Distill ($T=3$, KTO) & ~~~.850 & ~~.571 & ~~\textbf{.939}\textsuperscript{†} & ~~.575\textsuperscript{†} \\
\hline
\end{tabular}
\caption{Evaluation results for the WMT 2024 Metrics Shared Task. We employ SPA, ${\text{Acc}}_{\text{eq}}^{*}$, and $\textsc{SoftF1}$ as metrics at the system, sentence, and span levels, respectively. We report average scores across all translation directions. The best score is in bold. \textsuperscript{†} indicates significantly better performance than the best baseline method within the baseline group on all translation directions ($p<0.05$). xCOMET. *-Reg and *-ESD denote regression (score-only) and token-classification modes, respectively.
xCOMET-QE-*  indicates a reference-free variant; others are reference-based.}
\label{tab:main}
\end{table*}

\subsection{Experimental Setup}
\label{sec:exp_setup}

\paragraph{Datasets and Model}
We used all source and translation pairs $(s, t)$ from the WMT20–23 Metrics Shared Task \cite{wmtm20,wmtm21,wmtm22,wmtm23} datasets as $\mathcal{D}_{u}$. This ensures a fair comparison with Gold-* because both settings use identical $(s, t)$ pairs and differ only in their error-annotation training targets.
We used the MQM annotations from the WMT24 Metrics Shared Task \cite{wmtm24} as our test set, which covers three translation directions: English$\rightarrow$German, English$\rightarrow$Spanish, and Japanese$\rightarrow$Chinese. 
We used \textit{Qwen3-30B-A3B-Instruct-2507} \cite{qwen3}, a popular instruction-tuned LLM, as the base ESD model $M_{\theta^{(0)}}$.
During testing, we employed a greedy search.
We employed a GEMBA-MQM-style prompt \cite{gemba_mqm}, supplemented by the disambiguation technique proposed by Juraska et al.~\shortcite{metricx25}, which involves generating additional context for spans with ambiguous surface forms (Figure~\ref{fig:prompt}). Following prior work, we retain the \texttt{category} and \texttt{sub\_category} fields, but do not use them in training-data construction or meta-evaluation because no existing meta-evaluation metric assesses span predictions at these granularities.
We used guided generation \cite{guided_gen} to ensure that each hypothesis adheres to the specified JSON format.

\begin{figure}[!t]
\centering
\begin{lstlisting}[
  basicstyle=\ttfamily\small\selectfont,
  breaklines=true,
  breakatwhitespace=true,
  aboveskip=0pt,
  belowskip=0pt,
  backgroundcolor=\color{gray!10},
  frame=single,
  framerule=0.4pt,
  framesep=4pt,
  rulecolor=\color{gray!40}
]
Based on the source segment and machine translation surrounded with triple backticks, identify error types in the translation and classify them. The categories of errors are: 
1. Accuracy (addition, mistranslation, omission, untranslated text, hallucination)
2. Fluency (character encoding, grammar, inconsistency, punctuation, whitespace,capitalization, register, spelling, unnatural flow)
3. Style (awkward)
4. Terminology (inappropriate for context, inconsistent use)
5. Locale convention (address, currency, date, measurement, name, number, telephone, time)
6. Non-translation (None sub-category)
7. Other (None sub-category)
8. No-error (None sub-category)
    Each error is classified as one of three categories: critical, major, and minor. Critical errors inhibit comprehension of the text. Major errors disrupt the flow, but what the text is trying to say is still understandable. Minor errors are technically errors, but do not disrupt the flow or hinder comprehension.
    
Please output in JSON format where each error span contains the following fields: span, severity, category, sub_category, and span_with_context (span_with_context is optional. The shortest context text that can reliably eliminate ambiguity. Only include if the error span is not unique.)
If no errors are found, output an empty list `[]`
Please do not output anything else.
    
{source_lang} source:
    ```{source_seg}```
{target_lang} translation:
    ```{target_seg}```
\end{lstlisting}
\caption{Prompt template for the ESD models.}
\label{fig:prompt}
\end{figure}

\paragraph{Iterative MBR Distillation Configuration}
Due to computational resource constraints, we set $T=3$ and $C=256$.
Following Lyu et al.~\shortcite{esd-mbr}, we used $\textsc{SoftF1}$ as the MBR utility function for its robustness to empty annotations, and generated the candidate set via Top-$K$ ($K=10$) sampling at $temperature=2.0$. Following standard MBR practice, we did not deduplicate the candidates; ties in MBR score were resolved using model-estimated probabilities when selecting $E^+$ and $E^-$.

\paragraph{Baselines}
We compare against:
\begin{itemize}
  \item \textbf{xCOMET:} The state-of-the-art evaluation metric \cite{xcomet}. We evaluate its largest \textit{XXL} version in both reference-based (xCOMET) and reference-free (xCOMET-QE) settings. We report both its sentence-level regression scores (*-Reg) and token-level error span predictions (*-ESD).
  \item \textbf{Base Model:} direct prompting of $M_{\theta^{(0)}}$ with a GEMBA-MQM-style prompt.
  \item \textbf{Gold-*:} fine-tune $M_{\theta^{(0)}}$ on human annotations using SFT/DPO/KTO. The training data used was $\mathcal{D}_{u}$ with additional human annotations. For DPO and KTO, we use human annotations as $E^+$ and the output of $M_{\theta^{(0)}}$ with greedy decoding as $E^-$.
\end{itemize}

\paragraph{Training Details}
We randomly sampled a subset of data as the development set to select the checkpoint with the lowest training loss, while the remainder was used for model training, resulting in a development set of approximately $10$k examples and a training set of $193$k examples.
We performed full-parameter fine-tuning on the model using the AdamW optimizer \cite{adamw} with a learning rate of $1e-6$ and a warm-up step ratio of $0.05$. 
We set $\lambda=0.5$ for DPO and $\beta=0.5$ for KTO.
We set the batch size to $256$ (accounting for gradient accumulation and multi-GPU parallelism) and the maximum training epoch to $3$. 
During training, padding-free computations \cite{padding} and Flash Attention \cite{fa2} were used to accelerate training.
Our training codebase is based on MS-Swift \cite{ms-swift}, and training was completed on six machines, each equipped with 8$\times$ NVIDIA H200 GPUs.
Training of all models ran almost continuously on these machines for 10 weeks, with approximately 70\% of that time spent generating the large number of samples required for MBR decoding (we used vLLM \cite{vllm} to accelerate this process).

\paragraph{Evaluation}
Following prior work on ESD \cite{wmtm23,wmtm24,esd-mbr}, we used Soft Pairwise Accuracy (SPA) \cite{spa}, Pairwise Accuracy with Tie Calibration (${\text{Acc}}_{\text{eq}}^{*}$) \cite{acc_eq} and $\textsc{SoftF1}$ for system-, sentence- and span-level evaluation, respectively.
For SPA and ${\text{Acc}}_{\text{eq}}^{*}$, we employed the PERM-BOTH test \cite{perm_both}.
For $\textsc{SoftF1}$, we used paired bootstrap resampling \cite{pbs} for significance tests.
For each evaluation metric, we adopted the best baseline method within the baseline group as the baseline for significance tests. 
For example, Base Model serves as the baseline for SPA, while Gold-KTO serves as the baseline for ${\text{Acc}}_{\text{eq}}^{*}$.
For all ESD methods, system-level scores were calculated by averaging all sentence-level scores within that  MT system.

\subsection{Experimental Results}
\label{sec:results}

\paragraph{Main Result}
Table~\ref{tab:main} summarizes the overall performance.
In our reported results, the notation ``MBR Distill ($T$, training objective)'' indicates our proposed method trained for $T$ iterations using the specific training objectives (SFT, DPO, KTO).
MBR Distill ($T=2$, KTO) consistently improves over the Base Model and Gold-* baselines on SPA and $\textsc{SoftF1}$.
Furthermore, when compared to the state-of-the-art xCOMET models, our MBR Distill framework significantly outperforms both xCOMET-ESD and xCOMET-QE-ESD at the system and span levels. 
Thus, we conclude that our approach outperforms base model and human-annotated fine-tuned models at the system- and span-level.
However, on ${\text{Acc}}_{\text{eq}}^{*}$, MBR Distill performed only as well as the baselines.
Therefore, we conclude that our approach outperforms base model without ESD-specific and human-annotation-trained models at the system and span levels, while matching the performance of these baselines at the sentence-level.
This lack of sentence-level improvement may reflect the known difficulty of optimizing performance across evaluation granularities; Juraska et al.~\shortcite{metricx23} likewise report that simultaneous gains at the sentence and system levels remain challenging.
Given that our approach does not rely on any human-annotated ESD data, we believe it offers a novel paradigm for training ESD models.

\paragraph{Effect of Iterations $T$}
When $T$ is in $[1, 2]$, there is a strong positive correlation between $T$ and the scores on SPA, ${\text{Acc}}_{\text{eq}}^{*}$, and $\textsc{SoftF1}$, indicating the effectiveness of the iteration. 
However, when further increasing to $T=3$, the scores on these metrics actually decline compared to $T=2$. 
We analyze the reasons for this in Section \ref{sec:analysis_t}.

\paragraph{Effect of Different Training Objectives}
We did not observe any evidence that one training objective outperforms the others with Iterative MBR Distillation. 
Given that SFT is computationally cheaper (since it does not require a reference model $M_{\text{ref}}$), we recommend using SFT as the preferred training objective for Iterative MBR Distillation.

\begin{table*}[!h]
\centering
\small
\begin{tabular}{p{0.15\linewidth} | p{0.8\linewidth}}
\toprule
\textbf{Source} & The user interface is straightforward and easy to navigate. \\
\textbf{Translation} & \begin{CJK}{UTF8}{gbsn}用户界面是直接的并且很容易去导航。\end{CJK} \\
\midrule
\textbf{Human} & Span: ``\begin{CJK}{UTF8}{gbsn}直接的\end{CJK}'' (Minor) \\
& Span: ``\begin{CJK}{UTF8}{gbsn}去导航\end{CJK}'' (Minor) \\
\midrule
\textbf{Base Model} & Span: ``\begin{CJK}{UTF8}{gbsn}是直接的并且很容易去导航\end{CJK}'' (\textcolor{red}{\textbf{Major}}) \hfill \textit{$\leftarrow$ Over-penalizes translation style.} \\
\midrule
\textbf{Gold-SFT} & Span: ``\begin{CJK}{UTF8}{gbsn}直接的\end{CJK}'' (Minor) \\
& Span: ``\begin{CJK}{UTF8}{gbsn}去导航\end{CJK}'' (Minor) \\
\midrule
\textbf{MBR Distill}\newline($T=2$, KTO) & Span: ``\begin{CJK}{UTF8}{gbsn}直接的\end{CJK}'' (Minor) \\
& Span: ``\begin{CJK}{UTF8}{gbsn}去导航\end{CJK}'' (Minor) \hfill \textit{$\leftarrow$ Successfully calibrates severity without human data.}\\
\bottomrule
\end{tabular}
\caption{A qualitative example illustrating how the unadapted Base Model tends to over-predict error severity, grouping minor stylistic awkwardness (literal translation) into a single ``Major'' error. In contrast, both the human-trained Gold-SFT and our self-trained MBR Distill correctly identify the granular issues as ``Minor'' errors, aligning closely with human judgment.}
\label{tab:qual_example}
\end{table*}

\section{Analysis}
\label{sec:analysis}

\subsection{Analysis of $T=3$ Poor Performance}
\label{sec:analysis_t}
\begin{table}[H]
\centering
\begin{tabular}{lc}
\hline
\textbf{Method} & Utility Var\\
\hline
Base Model & $2.39 \times 10^{-2}$\\
\hline
MBR Distill ($T=1$, SFT) & $1.42 \times 10^{-2}$\\
MBR Distill ($T=2$, SFT) &  $1.69 \times 10^{-3}$\\
\cdashline{1-2}
MBR Distill ($T=1$, DPO) &  $4.86 \times 10^{-4}$\\
MBR Distill ($T=2$, DPO) &  $2.96 \times 10^{-4}$\\
\cdashline{1-2}
MBR Distill ($T=1$, KTO) &  $3.11 \times 10^{-3}$\\
MBR Distill ($T=2$, KTO) &  $1.57 \times 10^{-4}$\\
\hline
\end{tabular}
\caption{Estimated utility variance for different training configurations. The variance consistently decreases as the number of iterations ($T$) increases across all training objectives (SFT, DPO, KTO).}
\label{tab:var}
\end{table}
As mentioned in Section \ref{sec:results}, when $T$ increases from 2 to 3, the performance of MBR Distill stagnates or even declines. We hypothesize that this is because it becomes difficult to further decrease the MBR utility estimation error. 

To verify this hypothesis, following the theory of Kamigaito et al.~\shortcite{mbr_properties2}, we indirectly measure the utility error by calculating the variance of the estimated utility. The results are shown in Table \ref{tab:var}. 
We observe that for any variant of MBR Distill, as $T$ increases, the utility variance decreases progressively. 
This indicates that the utility estimation error struggles to decrease further as the iterations scale up. 
As a potential direction for improvement, we will explore methods to maintain the surface diversity of candidates during iterative training to indirectly alleviate the issue of declining utility variance in future work.

\subsection{Distribution and Severity of Error Spans}
\label{sec:analysis_number}
\begin{table}[!h]
\setlength\tabcolsep{6pt}
\centering
\small{
\begin{tabular}{lccc}
\toprule
\textbf{Method} & \textbf{Major} & \textbf{Minor} & $\frac{\textbf{Major}}{\textbf{Minor}}$\\
\midrule
Human & 9.0K & 15.5K & 0.58 \\
\midrule
Base Model & 16.4K & 16.1K & 1.02 \\
Gold-SFT & 5.4K & 11.8K & 0.46 \\
MBR Distill ($T=2$, KTO) & 13.7K & 21.9K & 0.63 \\
\bottomrule
\end{tabular}}
\caption{The total count and ratio of Major versus Minor error spans predicted by different models compared to human annotations on the WMT24 Metrics Shared Task test set.}
\label{tab:dist}
\end{table}
To better understand the behavior of the models, we investigate the distribution of error severities. 
Table~\ref{tab:dist} presents the total number of ``Major'' and ``Minor'' error spans, as well as their ratio, for human annotations, the untrained Base Model, the human-annotated Gold-SFT, and our best-performing iterative model, MBR Distill ($T=2$, KTO).

The results reveal that the unadapted Base Model heavily over-predicts the number of Major errors, resulting in a Major-to-Minor ratio (1.02) that is drastically higher than the human reference (0.58). 
In contrast, both the supervised Gold-SFT and our self-trained MBR Distillation effectively calibrate this distribution. 
Remarkably, without ever explicitly training on human annotations, MBR Distill ($T=2$, KTO) adjusts the model's severity recognition capabilities, achieving a Major-to-Minor ratio (0.63) that is highly aligned with human evaluation standards.

Counter-intuitively, this calibration occurs despite the fact that our chosen utility function (\textsc{SoftF1}) does not explicitly penalize the generation of excessive errors, nor does it inherently favor Minor errors over Major ones. 
This observation is consistent with the findings of Lyu et al.~\shortcite{esd-mbr}. 
It implies that the MBR distillation process intrinsically relies on the model's internal consensus to filter out overly harsh, hallucinatory evaluations (which are often idiosyncratic and lack consensus across the candidate set), thereby yielding a more natural and human-like severity distribution.

To provide an intuitive perspective on this phenomenon, Table~\ref{tab:qual_example} shows a qualitative example.
When faced with a translation that is grammatically correct but stylistically awkward due to literal translation (e.g., translating ``straightforward'' to ``\begin{CJK}{UTF8}{gbsn}直接的\end{CJK}''), the Base Model tends to aggressively aggregate the text into a single ``Major'' error span. 
Conversely, MBR Distillation successfully learns to tone down this overly critical behavior, identifying the specific spans as ``Minor'' issues, mirroring the behavior of models trained strictly on human annotations.

\section{Conclusions and Future Work}
\label{sec:conclusions}
In this work, we challenged the traditional reliance on costly and subjective human annotations for training ESD models. 
We introduced Iterative MBR Distillation, a self-evolution framework that leverages MBR decoding to generate high-quality, synthetic training signals from an LLM's own predictions. 
Through extensive experiments evaluating SFT, DPO, and KTO training objectives, we demonstrated that our self-trained models outperform both base model without ESD-specific and human-annotated models at the system and span levels, while successfully matching their performance at the sentence level. 
Ultimately, our findings signify a promising new paradigm for developing highly accurate and scalable ESD models free from the constraints of human supervision.

Finally, our analysis of iteration scaling revealed a performance bottleneck at higher iterations ($T=3$), driven by a reduction in candidate diversity and utility variance. Addressing this bottleneck by maintaining candidate diversity during iterative training represents a key focus for our future work.

\section{Information About Use of AI Assistants}
We used ChatGPT and Gemini solely for translation and language polishing during manuscript preparation. All AI-assisted text was reviewed and revised by the authors, who take full responsibility for the final content.

\section*{Limitations}
While our proposed Iterative MBR Distillation framework demonstrates promising results for ESD, we acknowledge several limitations in our current study.

Firstly, there is a potential metric bias in our span-level evaluation. 
Specifically, we employ \textsc{SoftF1} as the utility function to drive the MBR decoding and distillation process, and subsequently use the exact same metric to evaluate the model's span-level performance. This alignment between the optimization target and the evaluation metric may inherently favor models trained via our framework. 
Such metric-overfitting concerns have been widely discussed in prior literature regarding MBR decoding in machine translation \cite{metric_bias}. However, given the current lack of more suitable or orthogonal metrics for span-level ESD evaluation, we rely on \textsc{SoftF1} for this study and leave the exploration or development of independent evaluation metrics to future work.

Secondly, the empirical validation of our method is currently restricted to a single base Large Language Model (\textit{Qwen3-30B-A3B-Instruct-2507}). 
This limitation is primarily dictated by the sheer computational expense associated with our framework. As detailed in Section~\ref{sec:exp_setup}, generating massive candidate sets for MBR scoring and conducting iterative full-parameter fine-tuning required extensive computational resources and time. 
Consequently, we were unable to scale our experiments to verify the generalizability of the proposed method across models of varying sizes or different architectural families. Future research with access to broader computational resources could investigate how this self-evolution framework performs across a wider spectrum of LLMs.

Finally, we only validated the effectiveness of our method on WMT24 \cite{wmtm24}. 
Our conclusions have not yet been tested on more diverse language pairs or more challenging datasets \cite{wmtm25}, nor have we evaluated the comparability of automatic metric scores across language pairs \cite{xqmeval}.

\section*{Acknowledgments}
This work was partially supported by JSPS KAKENHI Grant-in-Aid for Early-Career Scientists 25K21290.

\bibliography{custom}

\end{document}